\documentclass[conference]{IEEEtran}
\IEEEoverridecommandlockouts
\usepackage{cite}
\usepackage{amsmath,amssymb,amsfonts}
\usepackage{algorithmic}
\usepackage{graphicx}
\usepackage{textcomp}
\usepackage{xcolor}
\usepackage{hyperref}
\usepackage{bm}
\usepackage{mathtools}
\usepackage{subcaption}
\usepackage{float}
\usepackage{geometry}
\usepackage{soul}

\def\BibTeX{{\rm B\kern-.05em{\sc i\kern-.025em b}\kern-.08em
    T\kern-.1667em\lower.7ex\hbox{E}\kern-.125emX}}

\begin{document}

% \title{Image-based Visual Servoing of an Aerial Robot Using Deep Learning Feature Extraction\\
\title{Deep Visual Servoing of an Aerial Robot Using Keypoint Feature Extraction\\
{\footnotesize}
\thanks{This work was supported by the National Research Council Canada, grant number AI4L-128-1, Natural Sciences and Engineering Research Council of Canada, grant number 2023-05542, and Natural Sciences and Engineering Research Council of Canada Discovery Grant 2017-06764.}
}

\author{
\IEEEauthorblockN{Shayan Sepahvand}
    \IEEEauthorblockA{Department of Mechanical, Industrial,\\
    and Mechatronics Engineering\\
    Toronto Metropolitan University\\
    Toronto, Canada\\
    Email: shayan.sepahvand@torontomu.ca}
\and
\IEEEauthorblockN{Niloufar Amiri}
    \IEEEauthorblockA{Department of Mechanical, Industrial,\\
    and Mechatronics Engineering\\
    Toronto Metropolitan University\\
    Toronto, Canada\\
    Email: niloufar.amiri@torontomu.ca}
\and
\IEEEauthorblockN{Farrokh Janabi-Sharifi}
    \IEEEauthorblockA{Department of Mechanical, Industrial,\\
    and Mechatronics Engineering\\
    Toronto Metropolitan University\\
    Toronto, Canada\\
    Email: fsharifi@torontomu.ca}
}

\maketitle

\begin{abstract}
The problem of image-based visual servoing (IBVS) of an aerial robot using deep-learning-based keypoint detection is addressed in this article. A monocular RGB camera mounted on the platform is utilized to collect the visual data. A convolutional neural network (CNN) is then employed to extract the features serving as the visual data for the servoing task. This paper contributes to the field by circumventing not only the challenge stemming from the need for man-made marker detection in conventional visual servoing techniques, but also enhancing the robustness against undesirable factors including occlusion, varying illumination, clutter, and background changes, thereby broadening the applicability of perception-guided motion control tasks in aerial robots. Additionally, extensive physics-based ROS Gazebo simulations are conducted to assess the effectiveness of this method, in contrast to many existing studies that rely solely on physics-less simulations. A demonstration video is available at \texttt{https://youtu.be/Dd2Her8Ly-E}. 

% Finally, a comparison is made between IBVS performed using man-made objects and an actual object from the dataset.
\end{abstract}

\begin{IEEEkeywords}
Image-based Visual Servoing (IBVS), Convolutional Neural Networks (CNNs), Robust Control, Aerial Robots.  
\end{IEEEkeywords}

\section{Introduction}\label{s1}

% The problem of the pose estimation of the objects in a real-world environment is a highly challenging tasks. It serves as a prerequisite to achieve motion control tasks. When it comes to the lab indoor environment, the precise and mostly expensive motion capture instruments provide the pose feedback of the objects. On the other hand, in outdoor scenarios, the position and orientation are measured utilizing a global positioning system (GPS) and an inertial measurement unit (IMU), respectively. As a result, in a GPS-denied environment sensors falls short to perform properly leading to instability in the position control loop. 

The pose estimation of objects is a critical challenge in real-world position-based motion control of aerial robots. In indoor laboratory environments, accurate pose feedback is provided by precise and expensive motion capture systems. In outdoor environments, as vision-based sensors are less effective when the captured image lacks sufficient texture, sensory feedback is typically provided by a global positioning system (GPS) and an inertial measurement unit (IMU). However, in GPS-denied environments, these sensors fail to perform reliably, leading to instability in the position control loop.

% A cost-effective alternative involves utilizing visual data from cameras, which has driven researchers to develop diverse visual servoing techniques, generally categorized as image-based visual servoing (IBVS) and position-based visual servoing (PBVS), and hybrid VS. In robotics, numerous applications rely on visual data to derive control signals for robotic platforms including tasks such as aerial manipulation \cite{Amiri2024}, grasping \cite{Albert2024}, moving target tracking \cite{Kumar2024, Sepahvand2024}, and autonomous quadrotor landing \cite{Mu2025,Sepahvand2024Landing}. 

A cost-effective and widely accessible alternative involves utilizing visual data from cameras, which has driven researchers to develop diverse visual servoing techniques, generally categorized as image-based visual servoing (IBVS), position-based visual servoing (PBVS), and a combination of both. In the context of robotic systems, numerous applications rely on visual data to generate control signals for robotic platforms, including tasks such as aerial manipulation \cite{Amiri2024}, grasping \cite{Albert2024}, moving target tracking \cite{Kumar2024, Sepahvand2024}, and autonomous quadrotor landing \cite{Mu2025, Sepahvand2024Landing}. 

% Marullo et al. \cite{Marullo2023} categorized the pose estimation visual servoing approaches as template-based, feature-based, and learning-based methods. In an instance of time, it is desired to estimate the pose of an object in real-time to enable the outer-loop motion controller substantiating the significance of monocular single-shot 6-D object pose estimation. These algorithms incorporate depth or RGB data to achieve their tasks \cite{Thalhammer2024}. There are currently datasets of real-world images that may be utilized to test the effectiveness and reproducibility of different approaches. The major attributes of these datasets is the variety of the objects they contain, number of images, and the characteristics such as varying illumination levels \cite{xiang2017}, strong object occlusion \cite{Brachmann2014}, symmetries \cite{wang2019}, clutter \cite{chen2022}, multiple object instances per image \cite{wang2022phocal}, lack of texture\cite{he2023}, object transparency \cite{liu2020}. Apart from the dataset and the model used for training, in the context of PBVS for aerial robots, the forward propagation to generate the pose from the incoming visual data stream and the continuous tracking of the target are more pronounced compared to quasi-static applications.

Marullo et al. \cite{Marullo2023} categorized pose estimation approaches into template-based, feature-based, and learning-based methods. At any given moment, it is desirable to estimate an object's pose relative to the aerial platform in real-time to enable the outer-loop motion controller, using only monocular single-shot 6D object pose estimation. These algorithms incorporate depth or RGB data to facilitate the pose estimation task \cite{Thalhammer2024}. Currently, there are datasets of real-world images that can be used to evaluate the effectiveness and reproducibility of different approaches in aerial applications. The key attributes of these datasets include the variety of objects they contain, the number of images, and specific characteristics such as varying illumination levels \cite{xiang2017}, strong occlusion \cite{Brachmann2014}, symmetries \cite{wang2019}, clutter \cite{chen2022}, multiple object instances per image \cite{wang2022phocal}, lack of texture \cite{he2023}, and transparency \cite{liu2020}. Apart from the dataset and the model used for training, the time complexity of these models interferes with the real-time operation of the aerial platform. Real-time performance is essential for forward propagation to generate object poses from the incoming visual data stream and for continuous target tracking. This issue arises because most available models are trained for quasi-static applications, which makes them less suitable for dynamic aerial tasks.

% The IBVS focuses on the image space rather than the pose of the target. The common features are the SIFT corner points, image moments, lines, and ovals. Consequently, the feature errors are obtained  uses an interaction matrix (a.k.a. image Jacobian) to find the camera's local Cartesian velocity. Furthermore, depending on the configuration of the camera, two eye-in-hand (or end-point closed-loop) and eye-off-hand (end-point open-loop) groups are available. Despite issues such as out-of-field-of-view dysfunction and limited camera excursion, the method is inherently robust against depth variation and exhibits less computational burden compared to PBVS \cite{Corke2023}. 

In IBVS, the focus is on the coordinates of the features in the image space rather than the pose of the target. Common features include scale-invariant feature transform (SIFT) corner points, image moments, lines, and ovals. The feature errors are then computed using an interaction matrix (image Jacobian) to estimate the camera's local Cartesian velocity. Depending on the camera configuration, two types of camera setups, known as eye-in-hand (or end-point closed-loop) and eye-off-hand (end-point open-loop), are available. Despite challenges such as out-of-field-of-view issues and limited camera excursion, IBVS is inherently robust against depth variation and camera calibration errors. It also exhibits a lower computational burden compared to PBVS \cite{Corke2023}.

% The learning-based IBVS also has indicated different variations in the recent years. It generally can be categorized as deep direct and indirect techniques with two variations of processing the entire image or just the features. In \cite{Yu2019, Tokuda2021}, CNNs were applied to whole images, either to directly estimate the pose difference between the target and current image or to determine the robot’s control commands in an end-to-end, meaning that the model indirectly estimate the interaction matrix. Conversely, a feature-based approach benefits from an attention-like mechanism: as long as a sufficient number of features can be matched between the target and current images, Visual Servoing remains feasible even in the presence of significant distractions \cite{Adrian2022}. Even though the use of trained feature extraction CNNs may result in an enhanced accuracy, it may pose heavy computational bourdon on the edge devices of drones due to feature extraction, feature matching, outlier rejection steps. In our previous work \cite{Amiri2024Case}, we proposed a CNN predicting only a few points to perform the IBVS tasks for a rigid non-floating manipulator. This accordingly motivates us further robustness and efficiency of this approach for aerial robots under various undesirable conditions.

% \newgeometry{top=0.75in, bottom=0.75in, left=0.75in, right=0.75in}

Learning-based IBVS has also opened up new opportunities in image-based control of aerial platforms in recent years. This approach is generally categorized into deep direct and indirect techniques, with two sub-variations: one processing the entire image and the other processing only the features. In \cite{Yu2019, Tokuda2021}, the entire image is processed by the CNN, either to directly estimate the pose difference between the target and current image or to determine the robot’s control commands in an end-to-end manner, meaning the model indirectly estimates the interaction matrix. Additionally, a learning-based approach benefits from an attention-like mechanism. This means that as long as a sufficient number of features can be matched between the target and current images, visual servoing remains feasible even in the presence of significant distractions \cite{Adrian2022}. Even though the use of trained feature extraction CNNs may result in enhanced accuracy, it may pose a heavy computational burden on the edge devices of drones due to feature extraction and matching, as well as the outlier rejection steps. In our previous work \cite{Amiri2024Case}, we proposed a CNN predicting only a few points to perform fast IBVS tasks for a rigid non-floating manipulator. This accordingly motivates further improvement of the robustness and efficiency of this approach for aerial robots under various undesirable conditions.

% 

% The contributions of this paper are as follows:

% \begin{itemize}
%     \item Development of a robust deep-learning-based IBVS capable of performing under occlusion, varying illumination, and clutter mitigating the need for man-made markers to stabilize the position loop of the aerial robot.
    
%     \item The creation of an annotated dataset later used for training and applying transfer learning and stunning concepts to enhance the regression task.
    
%     \item Assessment of theoretical statements through extensive simulations using ROS Gazebo, in contrast to many existing studies that rely solely on MATLAB-based simulations. 
    
% \end{itemize}

The contributions of this paper are as follows:
\begin{itemize}
    \item Development of a robust deep-learning-based IBVS for motion control of an unmanned aerial vehicle (UAV), eliminating the reliance on man-made markers for stabilizing the platform's position control loop.
    
    \item Implementation of a physics-based simulator in ROS Gazebo to evaluate the performance of the algorithm in a real-time environment.
    
    \item In-depth analysis of the system’s performance under image occlusion, varying illumination, and in cluttered environments, demonstrating promising results.

\end{itemize}

The rest of the paper is organized as follows. We elaborate on the IBVS control design in Section \ref{s2}. Afterwards, the steps of the deep feature extraction are described in Section \ref{s3}. Section \ref{s4} presents the close-to-real-world simulation comparative results, followed by the concluding discussion in Section \ref{s5}.

% list of content...
\section{IBVS Control Design}\label{s2}
The objective of the perception-guided control system is to drive the image feature error vector to zero within a finite time, ensuring both smooth and stable convergence of the features in the image plane and the UAV configuration in the spatial frame. Here, the corner pixel coordinates \linebreak $\bm{s} = \left(u_1, v_1, ..., u_N, v_N\right)^T \in \mathbb{R}^{2N} $ are considered as image features, and $\bm{s}_d$ denotes the desired image feature vector. Let the image feature error be defined as follows: 
\begin{equation}
\bm{e} := \bm{s}_d - \bm{s},
\label{feature_error}
\end{equation}

% \noindent where $\bm{s}:\mathbb{R}^3 \rightarrow \mathbb{P}^2$ is the camera projection mapping, $\bm{K} \in \mathbb{R}^{3 \times 3}$ is the camera intrinsics matrix, $\bm{p} \in \mathbb{R}^3$ is the fixed world point corresponding to the 2D pixel coordinate in the image plane, $\bm{\xi}$ and $\bm{\xi_d}$ are the fixed current and desired relative poses of the target with respect to the camera coordinate frame, respectively. 

\noindent Taking the time derivative of the image feature error \eqref{feature_error}, and noting that the second term on the right hand side is a constant, yields:

\begin{equation}
    \dot{\bm{e}} = \prescript{c}{}{\bm{L}(\bm{s}_{ni}}, \prescript{c}{}{\bm{Z}_i})\prescript{c}{}{\bm{v}_c},
    \label{feature_error_rate}
\end{equation}

\noindent where $\prescript{c}{}{\bm{L}} = \left(\bm{L}_1, \bm{L}_2, ..., \bm{L}_n \right)^T \in \mathbb{R}^{2N \times 6}$ is the interaction matrix, $\prescript{c}{}{\bm{Z}_i}$ is the depth of each world point from the camera optical center expressed in camera frame, with the assumption of no prominent radial and tangential distortions, $\bm{s}_n = \left(x_{ni}, y_{ni} \right)^T$ represents the normalized coordinates of i-th feature derived as follows: 

\begin{equation}
    x_{ni} = (u_i - c_x)/f_x,\hspace{0.5cm}  y_{ni} = (v_i - c_y)/f_y.
    \label{coordinates}
\end{equation}

\noindent As the above values are normally reported in pixels, one can neglect the value of effective pixel size. Thus,

\begin{equation}
\prescript{c}{}{\bm{L}}_i = 
\begin{bmatrix}
- \frac{1}{Z_i} & 0 & \frac{x_n}{Z_i} & x_n y_n & -(1 + x_n^2) & y_n \\
0 & -\frac{1}{Z_i} & \frac{y_n}{Z_i} & (1 + y_n^2) & -x_n y_n & -x_n
\end{bmatrix}.
\label{eq:image_jacobian}
\end{equation}

\noindent The parameters, $f_x$, $f_y$, $c_x$, and $c_y$ represents the focal length and the principal point coordinates measured during the calibration. According to \cite{Corke2023}, the feature error \eqref{feature_error_rate} can be stabilized using the following control law:

\begin{equation}
\prescript{c}{}{\bm{v}_c} = \lambda \prescript{c}{}{\bm{L}}^\dagger \bm{e},
\label{eq:control_law}
\end{equation}

\noindent therein, $\bm{v}_c=\left(v_x, v_y, v_z, \omega_x, \omega_y, \omega_z \right)^T$, and $\lambda$ is a control parameter that affects the convergence speed, $\prescript{c}{}{\bm{L}}^\dagger$ represents the pseudo-inverse of the interaction matrix. Considering a static offset between the camera and the drone base link, the twist calculated above should be expressed in the UAV's frame. As a result,

\begin{equation}
\prescript{b}{}{\bm{v}_c} = {\text{Adj}_{\prescript{b}{}T_c}}\prescript{c}{}{\bm{v}_c},
\label{eq:adj_transformation}
\end{equation}

\noindent where prescript $b$ shows that the vector is expressed in the drone's base link. The $\text{Adj}(.) \in \mathbb{R}^{6 \times 6}$ shows the adjoint representation of the static transformation between the child (camera) and parent (UAV's base) coordinate frames, $\prescript{b}{}{\bm{T}_c}$, as follows:

\begin{equation}
\text{Adj}_{\prescript{b}{}T_c} = 
\begin{bmatrix}
    \prescript{b}{}{\bm{R}_c} &\prescript{b}{}{\tilde{\bm{t}}_c}\prescript{b}{}{\bm{R}_c}\\
      \bm{0}_{3\times3}& \prescript{b}{}{\bm{R}_c}\\
\end{bmatrix},
\end{equation}

\noindent therein, $\prescript{b}{}{\bm{R}_c}$ and $\prescript{b}{}{\bm{t}}_c$ are the rotational and linear parts of $\prescript{b}{}{\bm{T}_c}$, respectively. 

It is worth highlighting that, in the case of an underactuated aerial platform, the channels related to pitch and roll cannot be directly controlled. Instead, one can rely on the concept of a virtual camera, which shares the same inertial pose as the original camera frame but has zero roll and pitch angles. Next, by decoupling the $\omega_x$ and $\omega_y$ channels, as they are determined by $v_x$ and $v_y$, $\prescript{b}{}{\bm{v}_c} \in \mathbb{R}^{6}$ is reduced to $\prescript{b}{}{\bm{v}_c} \in \mathbb{R}^{4}$. We refer readers to \cite{Lee2012, Sepahvand2024} for more details on the integration of IBVS into an underactuated platform.

\section{keypoint extraction using deep learning}\label{s3}

In this section, the details of the dataset creation, model training, and the techniques to enhance the accuracy of the model are elaborated. 

% The first step before training a CNN model is to create the dataset. The target object in this case is a tea bag with four corners. The tea bag is positioned in front of the robot within its field of view. Subsequently, 400 pictures were captured by the robot's camera, each depicting the tea bag in various positions. The images were captured by the left camera of the ZED mini (StereoLabs, Paris, France) stereo camera. The camera was calibrated before taking the images. The calibration parameters using algorithm \cite{fetic2012procedure} are listed in Table II.

% The images were labeled based on the four corners of the target object, with careful consideration given to the order of these corners. The top-left corner was considered the first label, and the bottom-left corner was the fourth label. A combination of image processing techniques was used to automatically annotate the images. The annotation process includes converting the images into grayscale and then applying a Canny edge detector with a specific threshold. Finally, the quadrilateral is detected by finding the maximum and minimum intersections of edges with parallel lines having a certain slope. Some of the annotated samples are shown in Fig. \ref{fig:annotate}. After partitioning the dataset into training and testing categories, it was downloaded as a zip file. The labels were stored in xlsx files containing the coordinates of the corners in a specific order. Further processing, including data augmentation, was performed subsequently.
\subsection{Dataset Generation}

\begin{figure}[t]
    \centering
    \includegraphics[scale  = 0.35, trim = {2cm, 2cm, 8cm, 2cm}, clip]{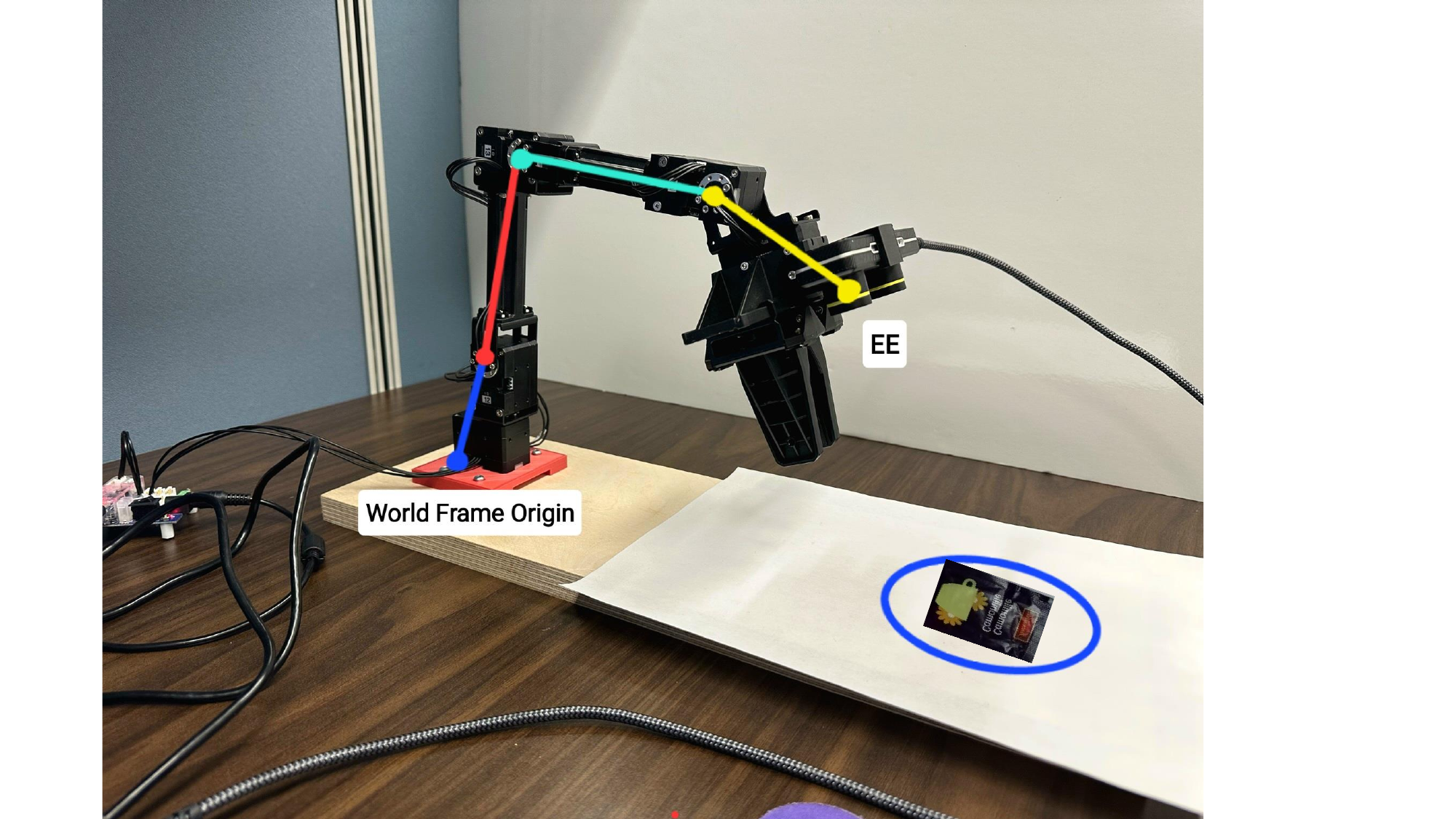}
    \caption{The mounted camera on the OpenMANIPULATOR-X with an eye-in-hand configuration for data collection}
    \label{real}
\end{figure}

\begin{figure}[t]
    \centering
    \begin{subfigure}[b]{0.2\textwidth}
        \centering
        \includegraphics[scale = 0.1, trim = {0cm 0cm 5cm 0cm}, clip]{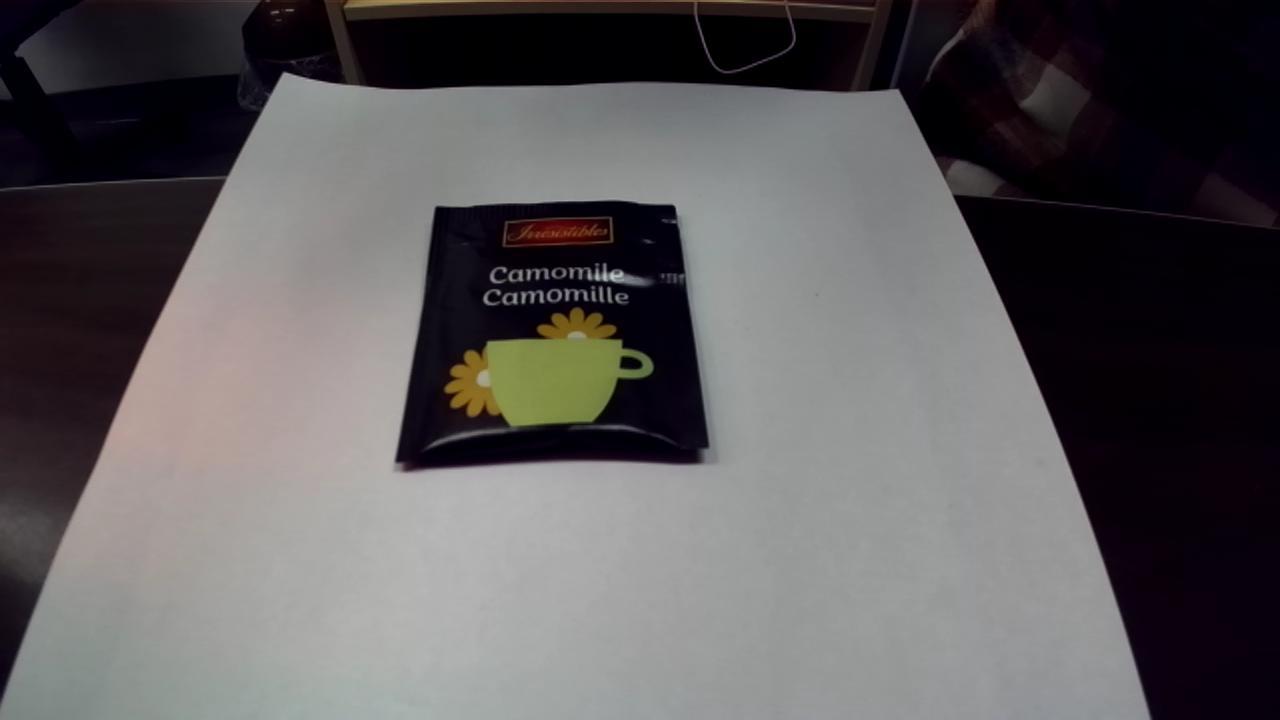}
        \caption{}
        \label{fig:image1}
    \end{subfigure}
    \hspace{0.5cm}
    \begin{subfigure}[b]{0.2\textwidth}
        \centering
        \includegraphics[scale = 0.1, trim = {0cm 0cm 5cm 0cm}, clip]{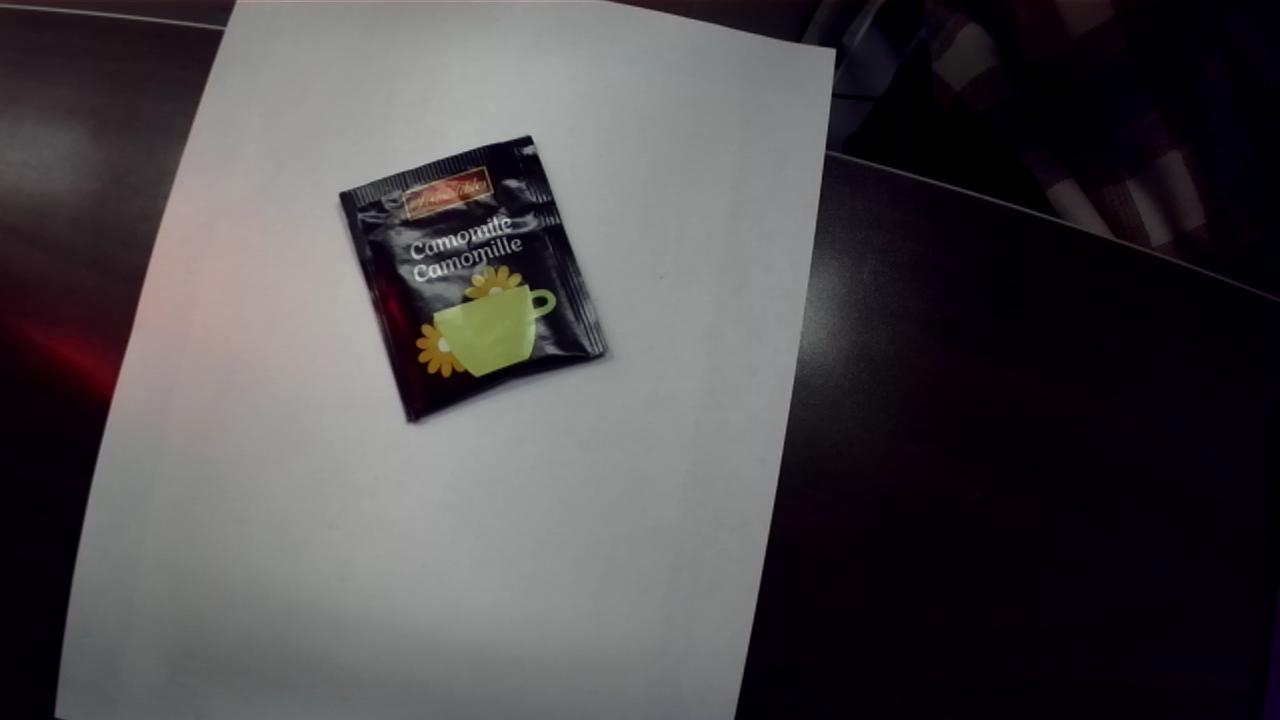}
        \caption{}
        \label{fig:image2}
    \end{subfigure}
    
    \vspace{0.2cm} % Add vertical space between rows if needed
    
    \begin{subfigure}[b]{0.2\textwidth}
        \centering
        \includegraphics[scale = 0.1, trim = {0cm 0cm 5cm 0cm}, clip]{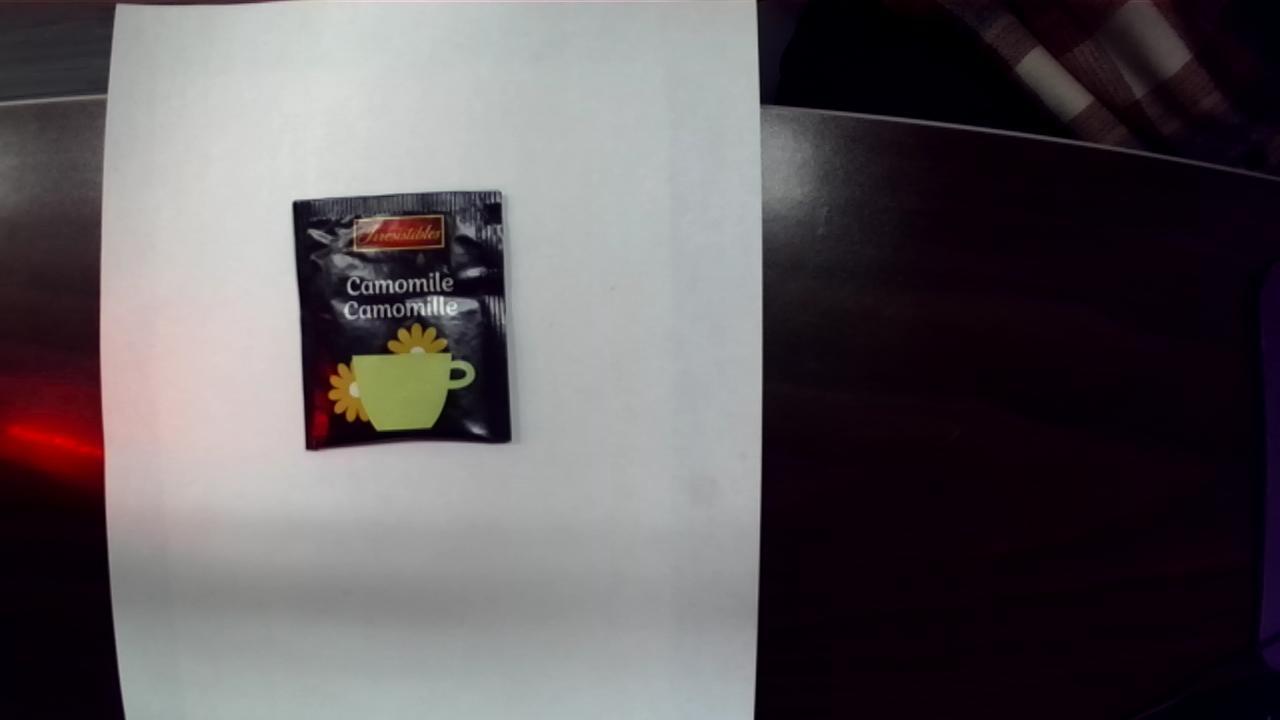}
        \caption{}
        \label{fig:image3}
    \end{subfigure}
    \hspace{0.5cm}
    \begin{subfigure}[b]{0.2\textwidth}
        \centering
        \includegraphics[scale = 0.1, trim = {0cm 0cm 5cm 0cm}, clip]{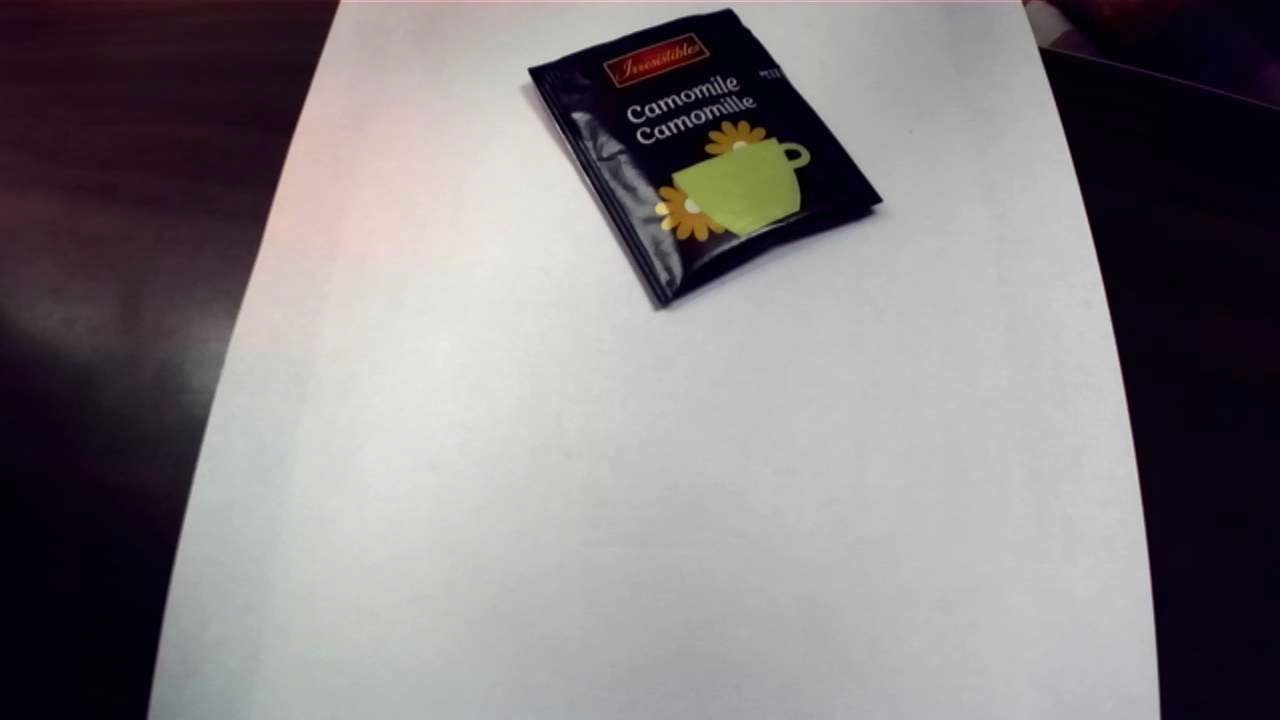}
        \caption{}
        \label{fig:image4}
    \end{subfigure}
    
    \caption{Sample of the images of the dataset.}
    \label{fig:2x2grid}
\end{figure}

The setup utilized for collecting the training data is shown in Fig. \ref{real}. The dataset created includes 400 RGB images captured using the ZED mini (StereoLabs, Paris, France) stereo camera (only one of the cameras is used) with an eye-in-hand configuration OpenMANIPULATOR-X. The object of interest is a rectangular teabag, placed in the field of view of the camera in an environment with a uniform background. Arbitrary, joint space values are commanded to capture images of the teabag in a wide range of camera viewpoints. The dataset size is quadrupled by rotating and flipping the original images. Examples of the captured images are displayed in Fig. \ref{fig:2x2grid}.

%  trim={<left> <lower> <right> <upper>}

The images were annotated by marking the four corners of the target object, with particular attention paid to the sequence of these corners. The top-left corner was designated as the first label, while the bottom-left corner was labeled as the fourth. A series of image processing methods were employed to automate the annotation process. This involved converting the images to grayscale and applying a Canny edge detector with a predefined threshold. The Canny edge detector is preferred due to its edge thinning and hysteresis thresholding capabilities. The quadrilateral shape was then identified by determining the maximum and minimum intersections of edges with parallel lines of a specific slope. 

\subsection{CNN Architecture}

The CNN architecture is composed of two sub-models. The primary model, VGG-19, is structured into five blocks, as outlined in \cite{Amiri2024Case}. Each block contains two or more convolutional layers designed for feature extraction, followed by a pooling layer to preserve the most important features. In the modified version, the traditional max-pooling layer is substituted with average-pooling to mitigate overfitting. The outputs are then flattened and fed into a fully connected layer, which supports the regression task and predicts eight pixel coordinates corresponding to four corners of each image.

\subsection{Training}

In this study, transfer learning methods were implemented within the TensorFlow framework to leverage pre-trained deep convolutional neural networks (CNNs) for feature extraction. Specifically, a VGG-19 model, pre-trained on the ImageNet dataset, was utilized. During the training process, the weights of the VGG-19 convolutional layers were frozen, while the fully connected dense layer was modified with a linear activation function, which is conventionally used in regression tasks. This modification was designed to optimize the chosen loss function, the Mean Absolute Error (MAE). The Adam optimizer was used to train the deep CNN over 300 epochs, with a batch size of 16 and an initial learning rate of \(10^{-5}\). The dataset was divided for training and validation with a split ratio of 0.1. 

\begin{figure}[htb]
    \centering
    \includegraphics[scale = 0.4, trim=0cm 0cm 0cm 2cm, clip]
    {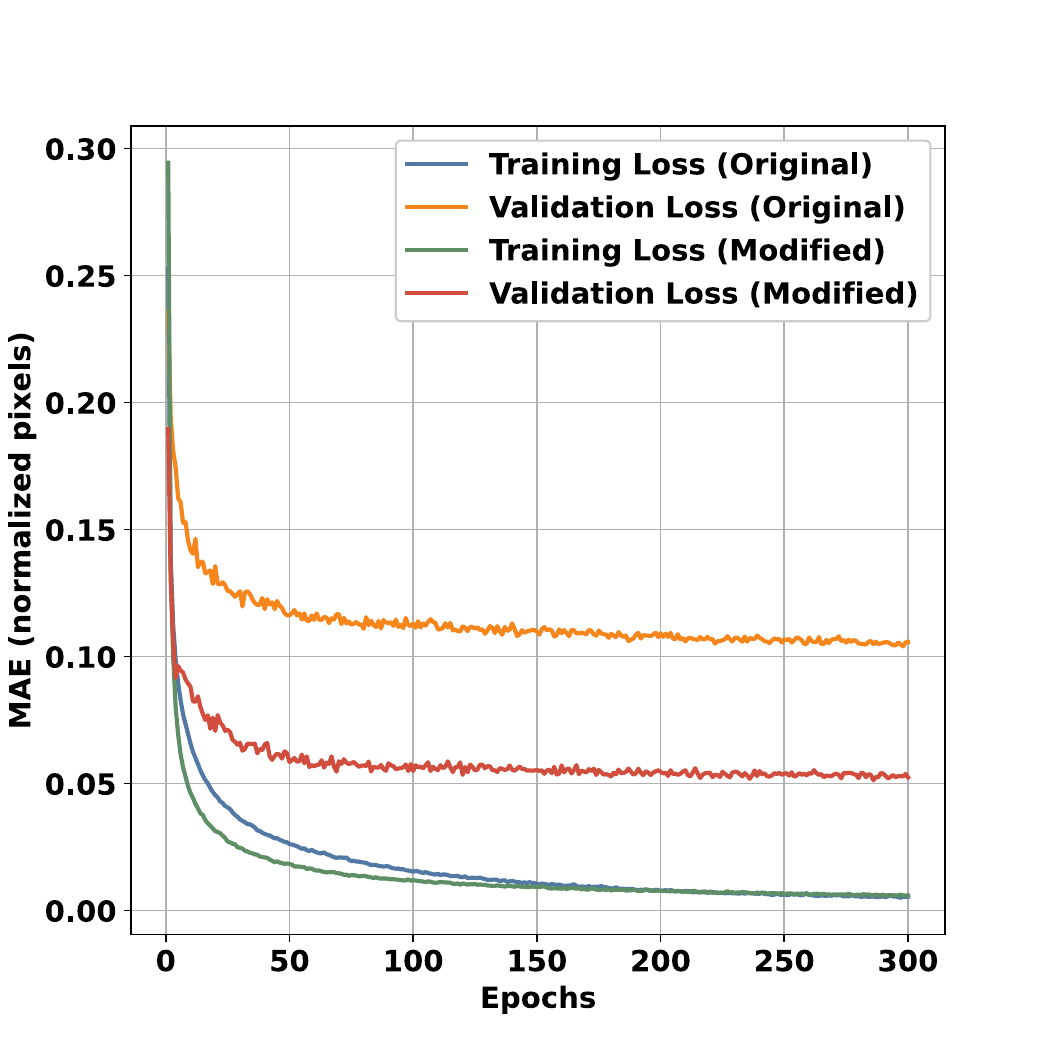}
    \caption{Learning curves showing how the training and validation loss functions change for two models.}
    \label{Learning Curve}
\end{figure}

Fig. \ref{Learning Curve} illustrates the learning curve for two variations of the model, namely the original and the modified versions, in terms of mean absolute error (MAE). In the latter, average pooling is used instead of max pooling. The output of the CNN consists of the normalized four corners. Here, normalization refers to dividing the pixel coordinates by the maximum spatial coordinate of the image. It can be observed that the MAE dropped from approximately 0.3 to about 0.007 pixels. Furthermore, the validation curve shows how the validation loss changes over epochs. Similarly, the validation loss decreased from 0.25 to just above 0.1.

\subsection{Implementation of the CNN for IBVS}
The CNN outputs the corner features of a custom object, which here is a teabag, and pass the output to the IBVS controller. During the training process, only four corners are considered. Thus, in \eqref{eq:image_jacobian}, the interaction matrix will have a size of $\prescript{c}{}{\bm{L}}  \in \mathbb{R}^{8 \times 6}$, and four pairs of the pixel coordinates $(u_i,v_i)$ in the \eqref{coordinates} are obtained by CNN.

%  trim=left bottom right top

\section{Experimental Results}\label{s4}

\begin{figure}[htbp]
    \centering
    
        % First row
        \begin{subfigure}[b]{0.22\textwidth}
            \centering
            \includegraphics[scale=0.065]{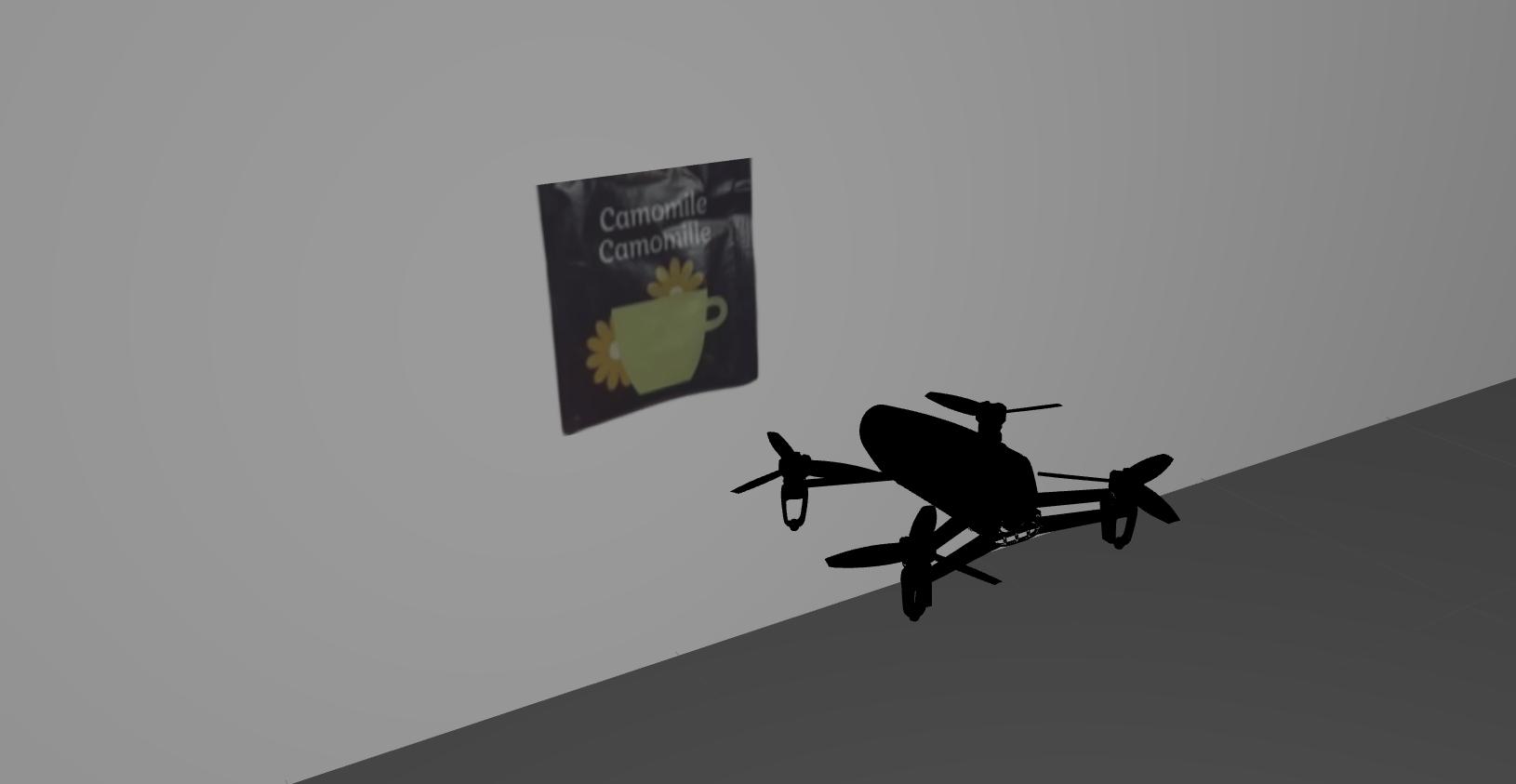} % Adjusted scale
            \caption{The Ideal Condition}
            \label{fig:sub_ideal}
        \end{subfigure} 
        \begin{subfigure}[b]{0.22\textwidth}
            \centering
            \includegraphics[scale=0.065]{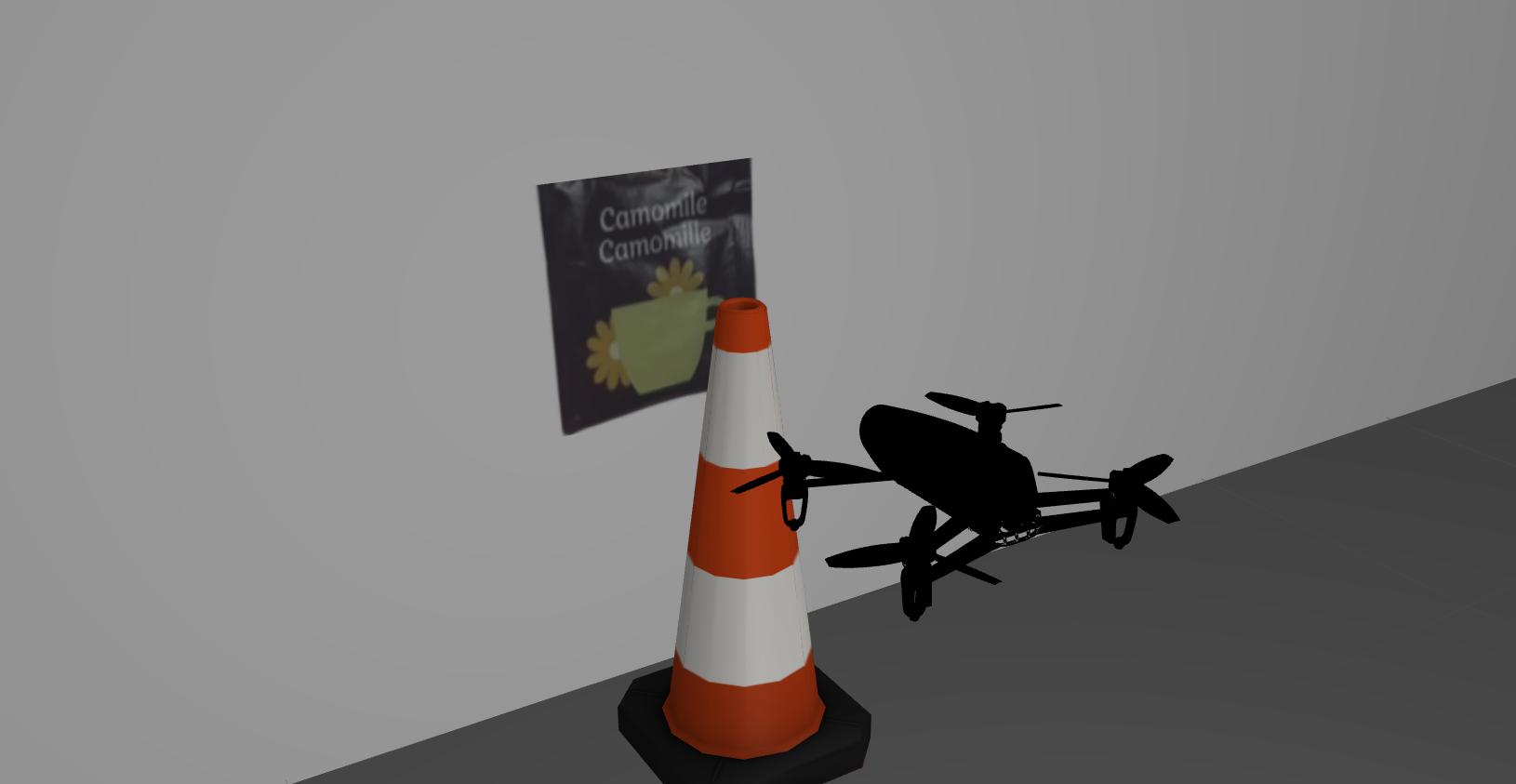} 
            \caption{Occluded Scene}
            \label{fig:sub_occluded}
        \end{subfigure} \\ 
        \vspace{0.5cm}
        % Second row
        \begin{subfigure}[b]{0.22\textwidth}
            \centering
            \includegraphics[scale=0.065]{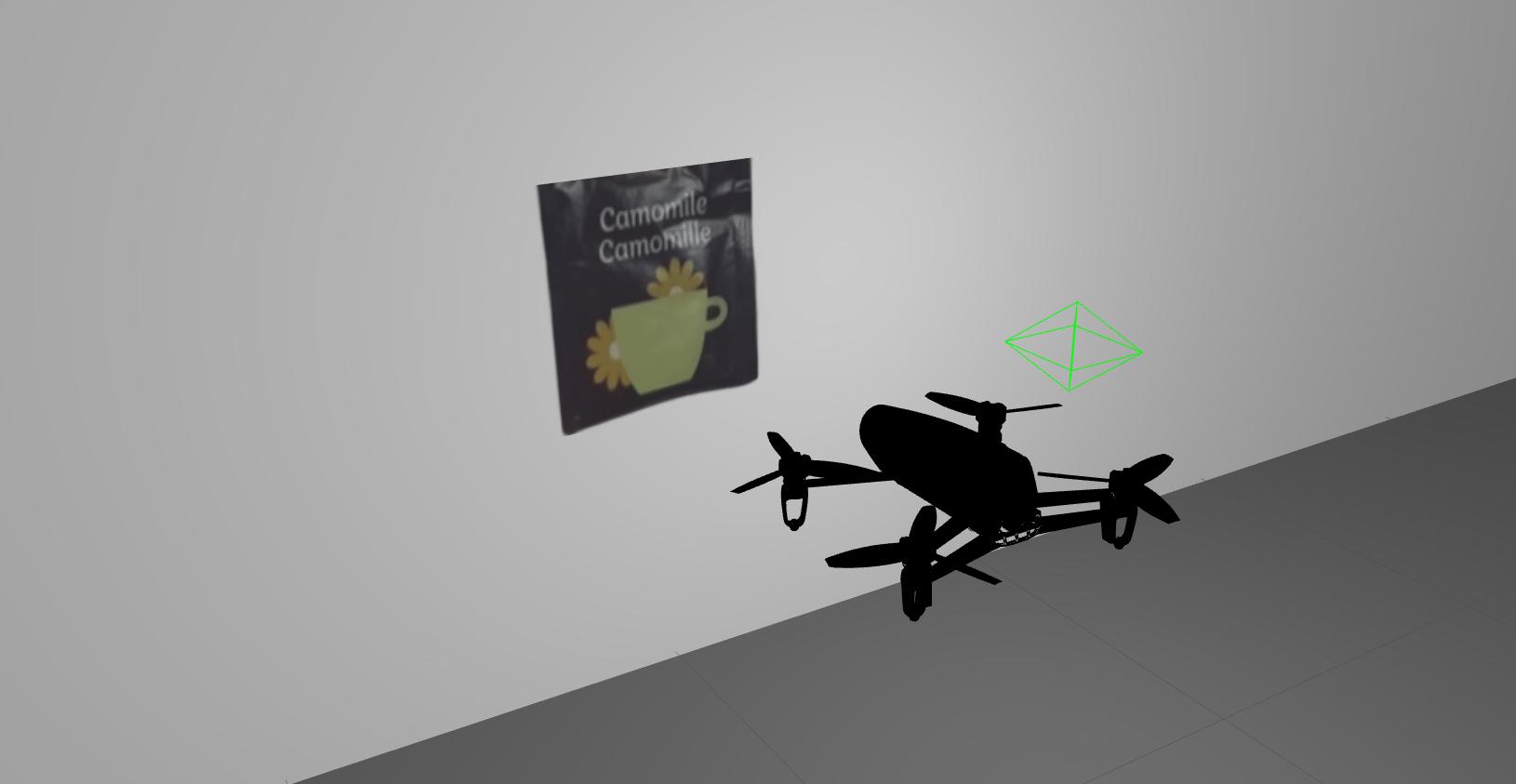} 
            \caption{Strong Illumination}
            \label{fig:sub_illumination}
        \end{subfigure} 
        \begin{subfigure}[b]{0.22\textwidth}
            \centering
            \includegraphics[scale=0.065]{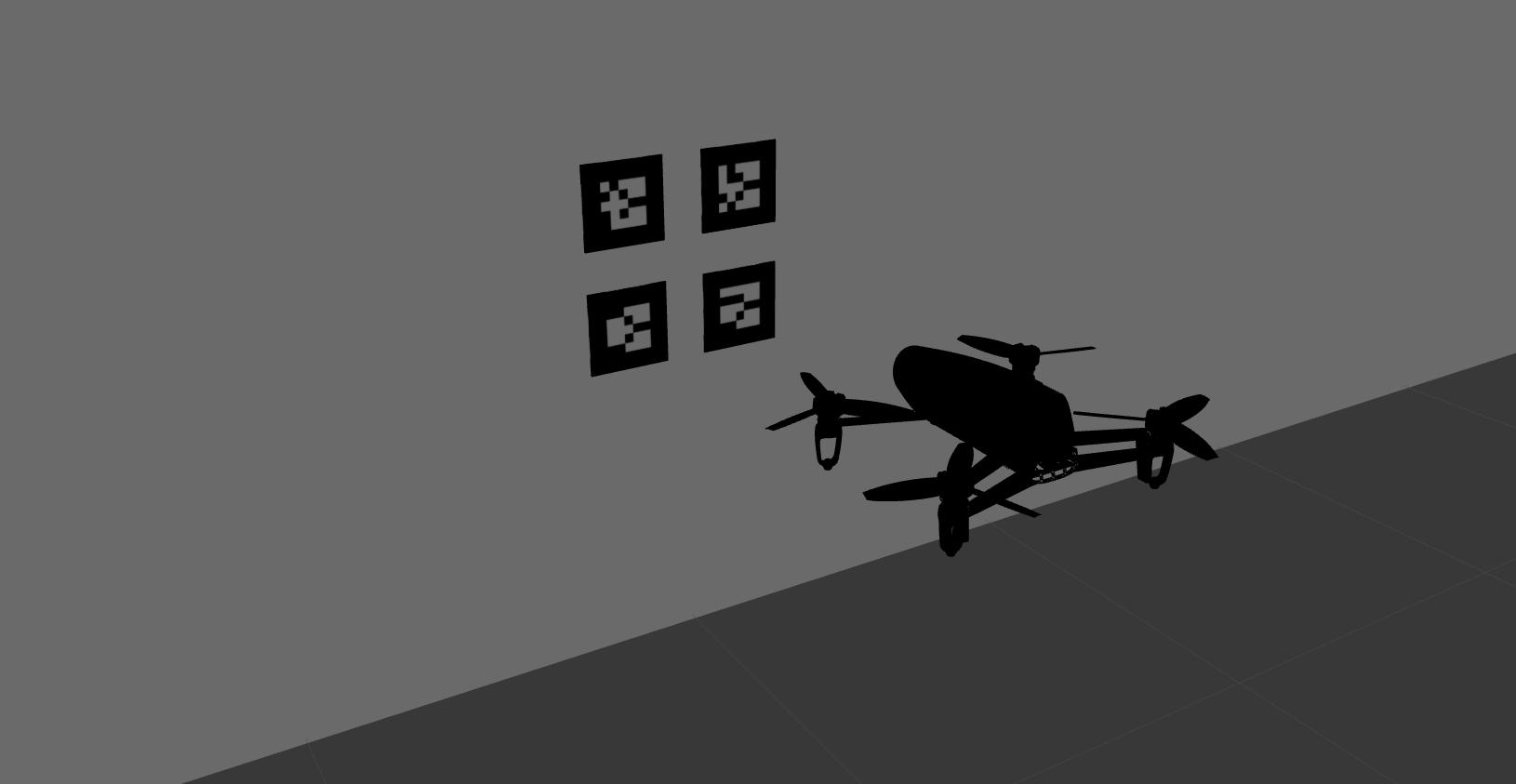} 
            \caption{Markers Used in \cite{Sepahvand2024}}
            \label{fig:sub_artags}
        \end{subfigure} \\ 
        \vspace{0.5cm}
        % Third row
        \begin{subfigure}[b]{0.22\textwidth}
            \centering
            \includegraphics[scale=0.068, trim={0cm 0cm 3cm 0cm},clip]{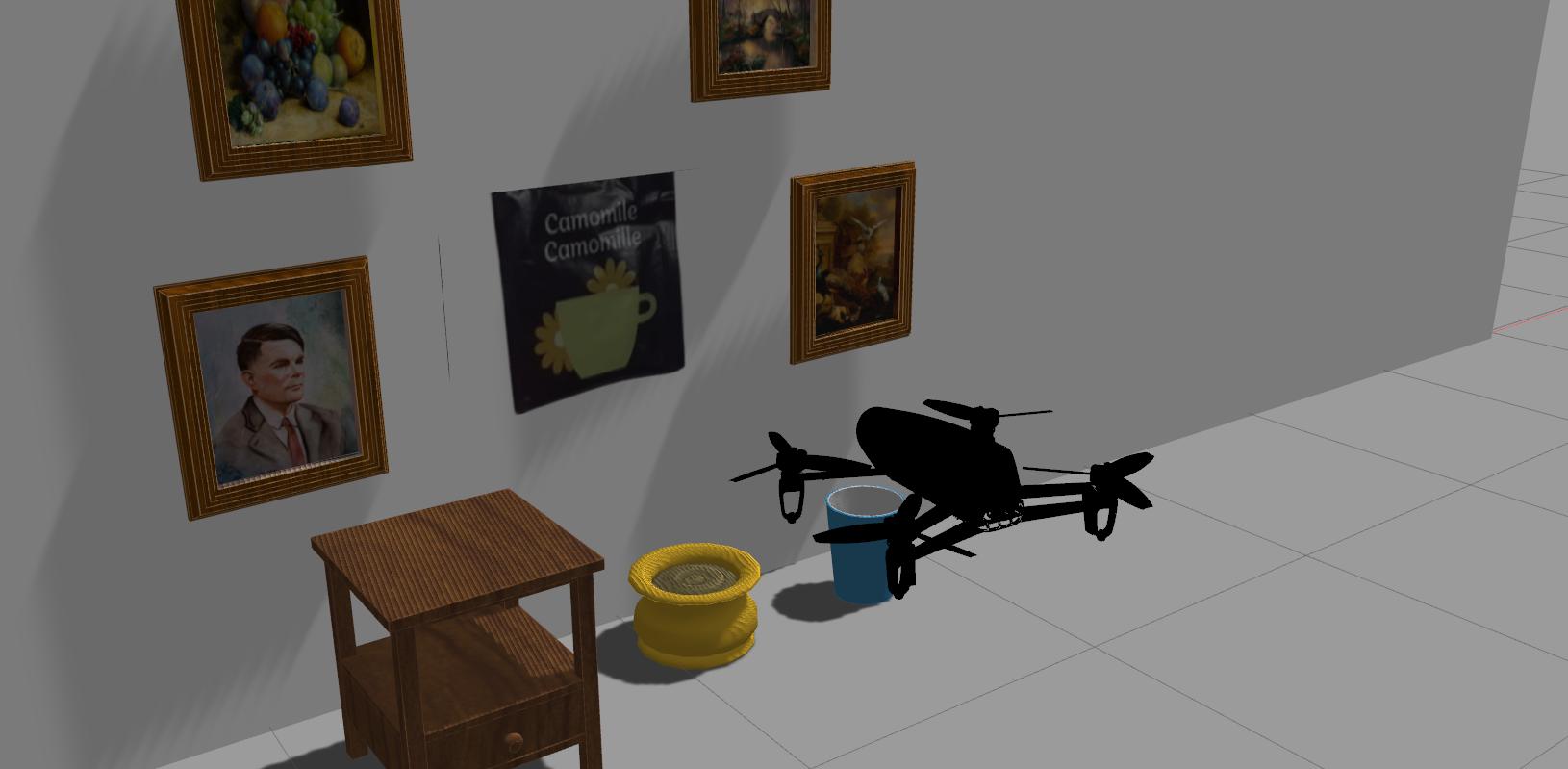} 
            \caption{Cluttered Environment}
            \label{fig:sub_clutter}
        \end{subfigure} 
        \begin{subfigure}[b]{0.22\textwidth}
            \centering
            \includegraphics[scale=0.065]{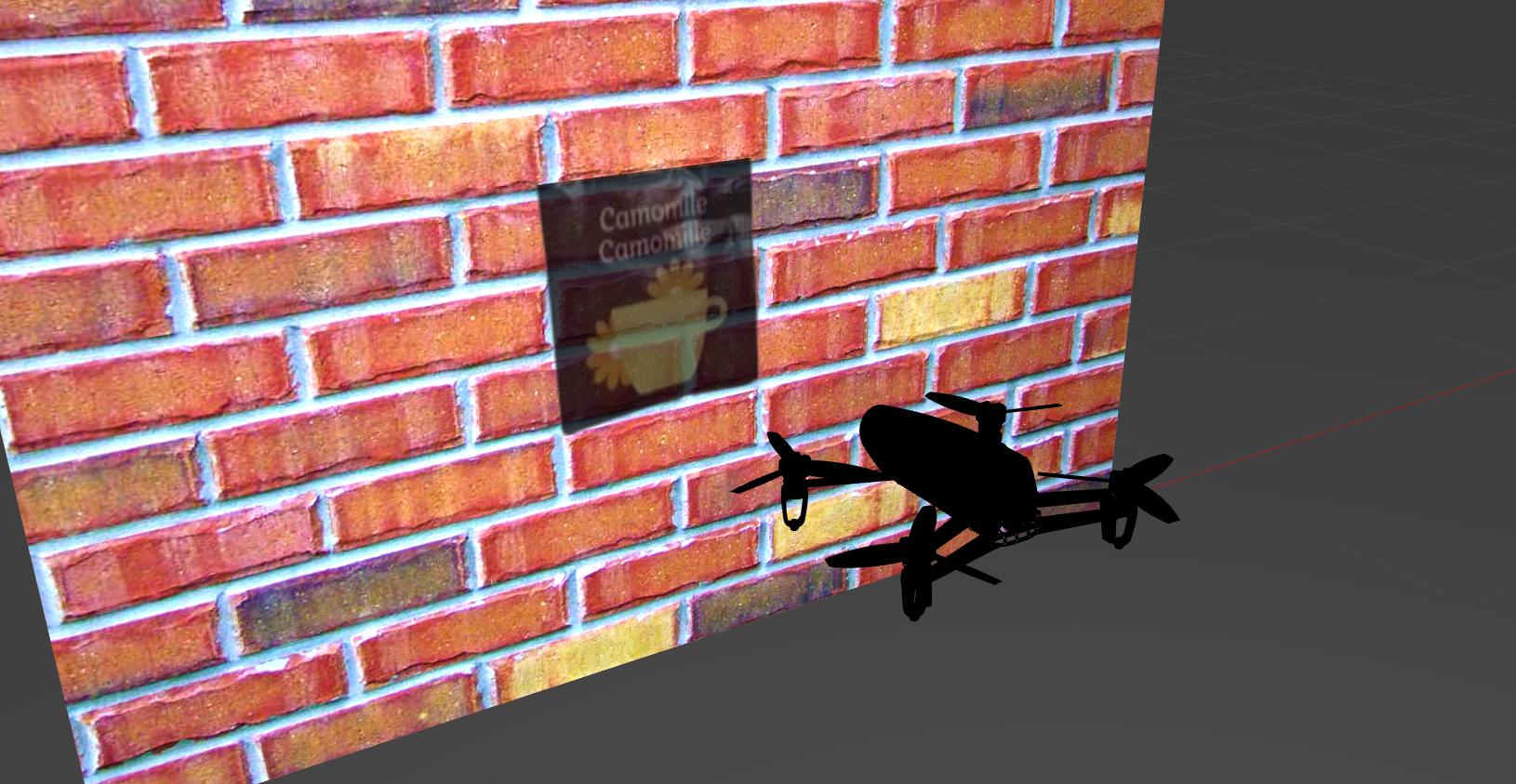} 
            \caption{Background Change}
            \label{fig:sub_background}
        \end{subfigure}

    \caption{Various worlds created in Gazebo were utilized to carry out robustness tests.}
    \label{fig:gazebo_worlds}
\end{figure}

To verify the effectiveness and robustness of the servoing algorithm in an indoor but partially unknown environment, five distinct worlds were spawned in the Gazebo\footnote{Gazebo simulation software: [online]. Available: \href{https://gazebosim.org/home}{https://gazebosim.org}, accessed on 20 Jan. 2024} simulator, as illustrated in Fig. \ref{fig:gazebo_worlds}. Each of these worlds represents scenarios occurring in real-world conditions, such as illumination variations, occlusions, background changes, and cluttered environments. For the sake of generalization, we also included the markers used in our previous works \cite{Sepahvand2024}. The Parrot Bebop 2 drone is simulated in Gazebo along with a built-in RGB camera featuring a full HD (1080p) resolution and an integrated 14-megapixel fisheye lens. The simulated quadrotor's specifications are listed in Table \ref{tab1}. ROS Noetic, installed on an Ubuntu 20.04 operating system, is utilized to develop the required nodes using C++14 and Python 3.8.10, processed by a 12th Gen Intel(R) Core(TM) i7-12700 CPU. The time interval for data collection is approximately $59$ seconds for all experiments. The control parameter $\lambda$ in the control law \eqref{eq:control_law} is set to 0.1. Increasing $\lambda$ reduces the convergence time but results in large camera twists, which may not be feasible in practice. As given in \eqref{eq:adj_transformation}, the control input is the UAV twist expressed in its local frame.

The time required for forward propagation of the image through the CNN ranges from $60$ to $130$ milliseconds. The images undergo histogram equalization to compensate for poor lighting conditions and are resized to match the input shape expected by the trained CNN. This preprocessing step takes approximately $90 \pm 20$ milliseconds. The controller node is significantly less computationally expensive, requiring only $2$ milliseconds to compute the quadrotor's body twist. Additionally, the image topic from the onboard camera publishes messages at a frequency of $30$ Hz.

\begin{figure*}
    \centering
    \includegraphics[scale = 0.4]{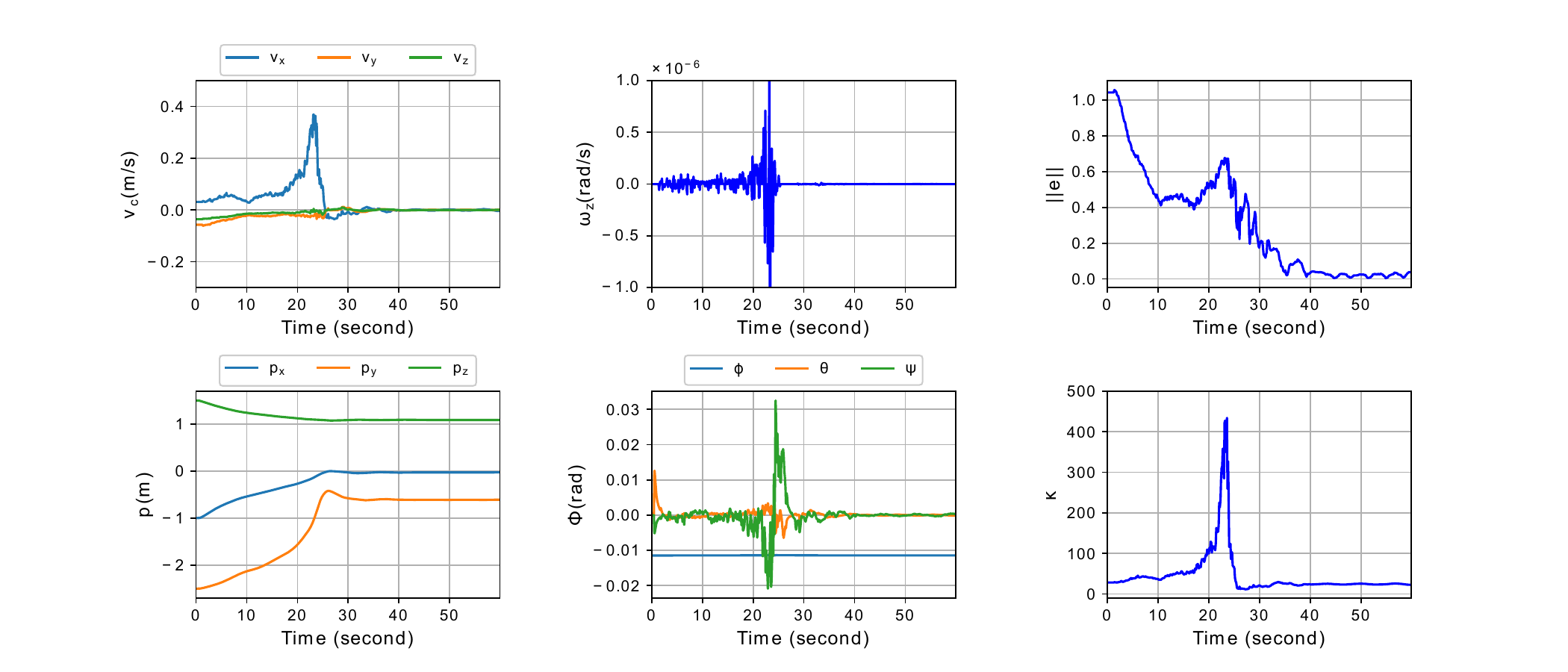}
    \caption{The performance of the controller in the absence of undesirable factors}
    \label{fig:normal}
\end{figure*}

\begin{figure*}
    \centering
    \includegraphics[scale = 0.4]{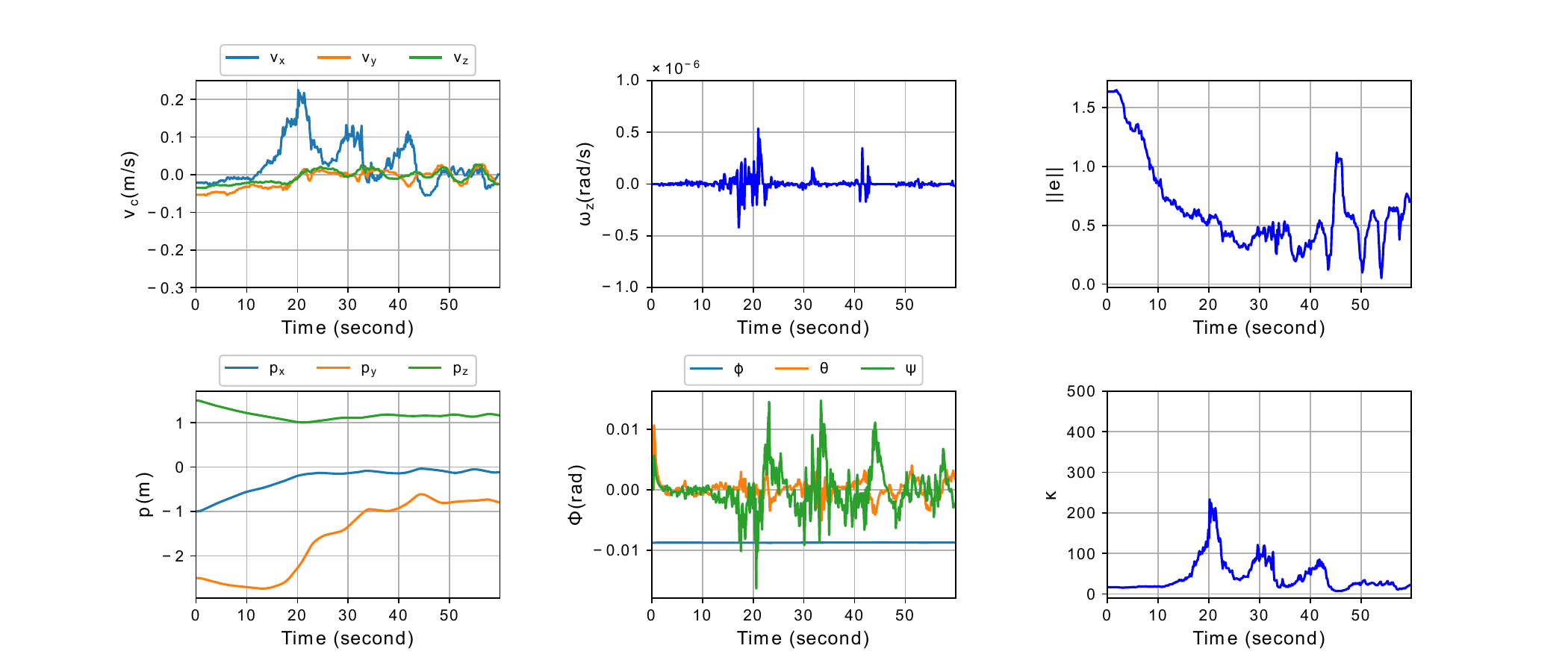}
    \caption{The kinematics of the quadrotor and the image when the scene is partially occluded}
    \label{fig:occlusion}
\end{figure*}

\begin{figure*}
    \centering
    \includegraphics[scale = 0.4]{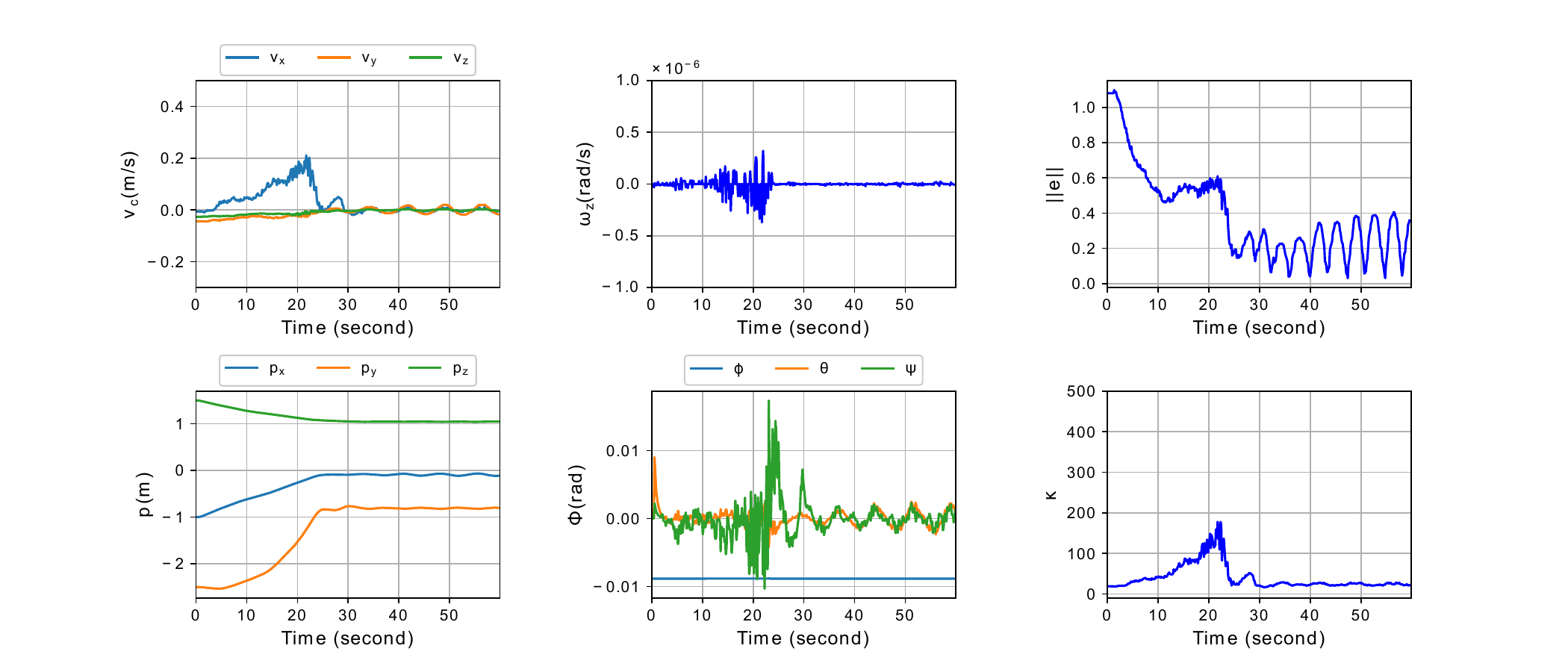}
    \caption{The generated signals when the illumination level is higher than the normal condition}
    \label{fig:illumination}
\end{figure*}

\begin{figure*}
    \centering
    \includegraphics[scale = 0.4]{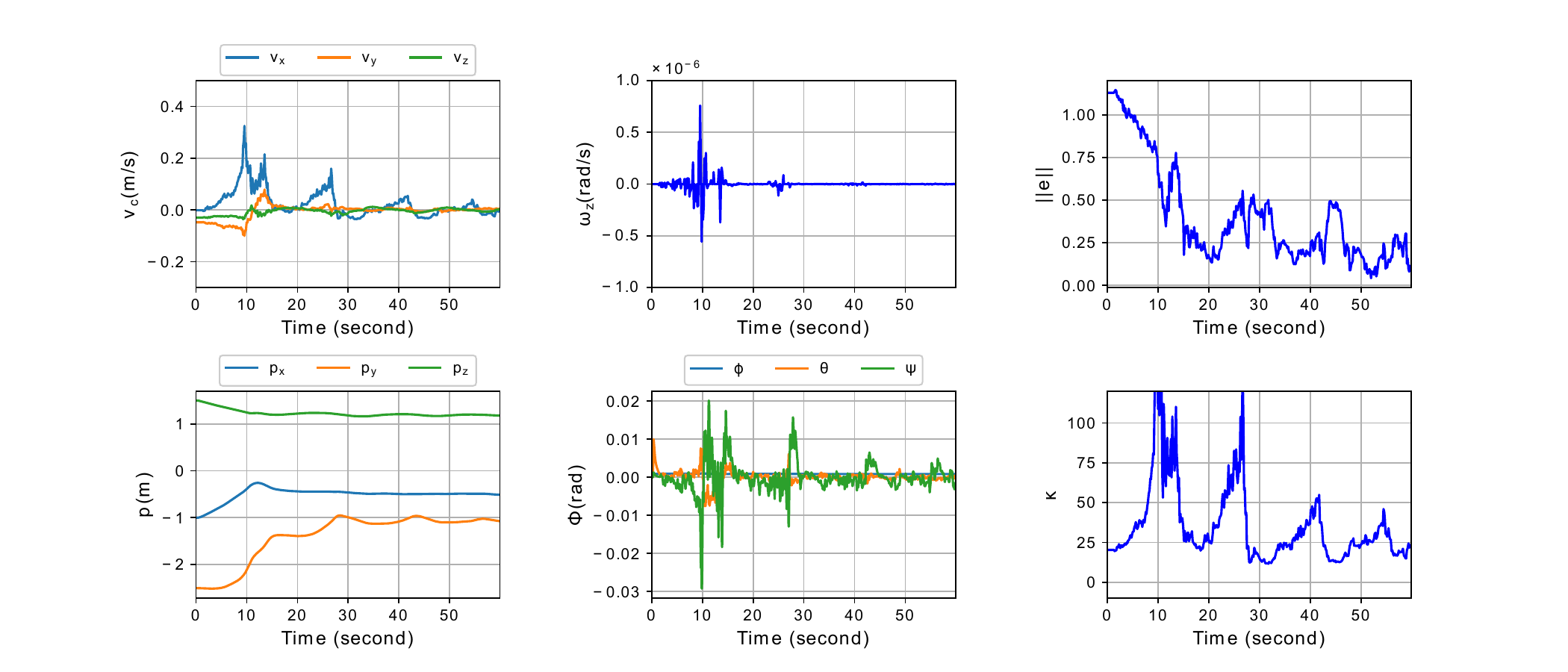}
    \caption{The effect of a cluttered environment on the closed-loop control system}
    \label{fig:clutter}
\end{figure*}

\begin{figure*}
    \centering
    \includegraphics[scale = 0.4]{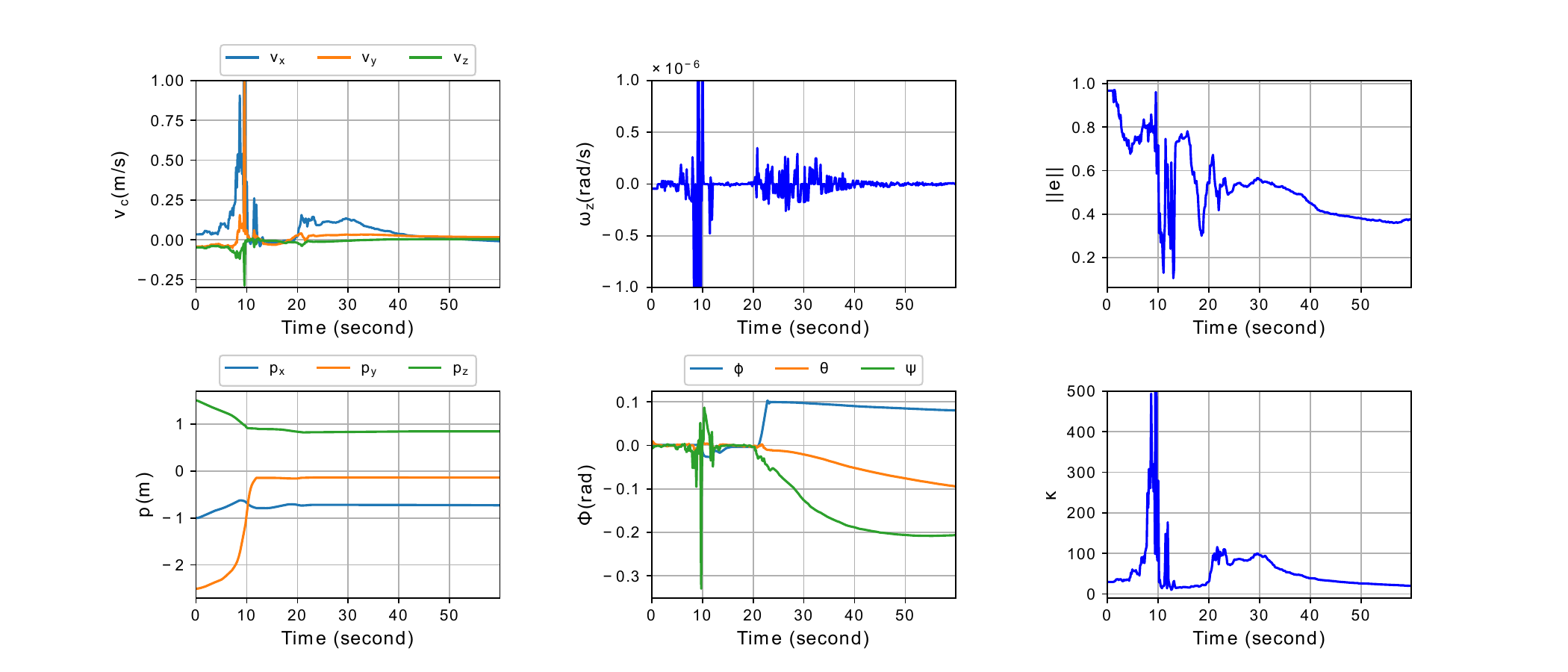}
    \caption{The response of the system to changes in the background}
    \label{fig:background}
\end{figure*}

Figures \ref{fig:normal}–\ref{fig:background} show the results of the experiments conducted for various scenarios, each of which is associated with an undesirable effect. In all of the aforementioned figures, the linear and angular Cartesian velocities of the camera, expressed in the local frame, are shown in the upper left and upper center sub-figures. Meanwhile, the position expressed in the global frame, $p=\left(p_x,p_y,p_z\right)^T$, and ZYX Euler angles, denoted with $\Phi=\left(\phi, \theta, \psi\right)^T$, of the quadrotor in the world (spatial) frame are displayed in the lower left and center images. The Euclidean norm of the normalized image feature error vector, $\|\bm{e}\|$ (dimensionless), shown in the upper right, serves as a criterion for evaluating the performance of the closed-loop system. Finally, the condition number of the interaction matrix is illustrated at the bottom left of the image, indicating near-singularity conditions.

The initial position of the quadrotor is \((-1, -2.5, 1.5)\, \mathrm{m}\) in the world reference frame, and the goal is to reach \((0, -0.8, 1)\, \mathrm{m}\). Regarding the orientation of the quadrotor, the pitch and roll angles are not directly controlled, as the platform is underactuated. However, we have aimed to maintain the same yaw angle of the UAV in both the final and initial poses. It is worth highlighting that the pose data of the UAV---and, accordingly, the camera---are not used in the error dynamics stabilization and are only reported to indicate the accuracy of the position controller.

Under ideal conditions, as Fig.~\ref{fig:normal} shows, the normalized image feature error is stabilized and approaches zero after $40$ seconds. The effect of the camera retreat is seen in the position and twist graphs, where the UAV approaches the target closer than the desired setpoint. This is accompanied by sharp changes in the condition number of the interaction matrix at approximately \(t = 23\) seconds. The occlusion test results, shown in Fig.~\ref{fig:occlusion}, indicate the robustness of the feature extraction method against partial occlusion of the corners. However, as a trade-off, it disturbs the detection of one of the corners, leading to oscillations in the commanded \(v_x\). The graphs shown in Fig.~\ref{fig:illumination} demonstrate how increasing the illumination has a disadvantageous effect on feature detection. Even though the performance of the controller is deteriorated compared to the ideal condition, it still reaches the desired final position with small fluctuations.

Fig. \ref{fig:clutter} shows the output of the algorithm in a cluttered environment. This scene is created by including several random images of the same size as the target image, as shown in Fig. \ref{fig:sub_clutter}. In this scenario, the features are correctly identified during the camera's expedition. However, at certain time instances, the predictions are incorrect, leading to sudden changes in the commanded twist. Nonetheless, this inaccuracy does not persist for a long duration, as the correct corners are predicted at multiple time instances following the false prediction. Lastly, a brick wall is used as the background for the target, and the norm of the image error indicates a steady-state error of $0.4$ when the background changes. As a result, one can deduce that the system is sensitive to background variations, leading to poor performance. This can be attributed to the fact that the training dataset does not account for unforeseen images with significant changes in pixel intensities. It can be observed that at $t = 22$ s, the yaw angle becomes nonzero, indicating that the UAV has crashed into the wall.
 
One of the challenges in conducting real-world experiments is the loss of feature detection during agile UAV maneuvers, which can significantly impact system performance. Moreover, uncertainties in camera intrinsic parameters introduce further complexities.

\begin{table}[ht]
\caption{Simulated Parrot Bebop drone and camera parameters.}
\label{tab1}
\centering
\renewcommand{\arraystretch}{1.2} % Adjust row spacing
\begin{tabular}{|l |p{3cm}| c|}
\hline
\textbf{Parameter} & \textbf{Value} & \textbf{Unit} \\
\hline
UAV Mass (with hull) & $0.399$ & $\mathrm{kg}$ \\
UAV Dimensions&$328 \times 328 \times 89$ & $\mathrm{mm}$ \\
UAV Moment of Inertia & $i_{xx}= i_{yy} = 0.01152$, $i_{zz} = 0.0218$
 & {$\mathrm{kg \cdot m^2}$} \\
Camera Focal Length & $462.137$ & pixel \\
Camera Principal Point & $(320, 240)$ & pixel \\
Radial/Tangential Distortion & neglected & -- \\
\hline
\end{tabular}
\end{table}

\vspace{-5cm}

\vspace{-5cm}

\vspace{-5cm}

\vspace{-5cm}

\vspace{20cm}
\section{Conclusions}\label{s5}

This brief studied the deep learning-based Image-Based Visual Servoing (IBVS) control of an aerial robot. First, a model was trained using a dataset created with a teabag as the target. These features were then used as alternatives to conventional corner points to generate the normalized image feature vector and the interaction matrix. The performance of the controller heavily depended on the feature extraction subsystem. Accordingly, four different real-world scenarios—where the scene was affected by illumination variations, occlusion, background changes, and clutter—were investigated in the ROS Gazebo simulation environment. The results were compared against ideal scenarios where only the target was present in the scene. The system demonstrated robustness against illumination variations, occlusion, and clutter, but struggled when the background changed. Cluttered and occluded environments negatively impacted performance, as false corners from other objects in the scene were detected, leading to oscillations in the commanded twist. Future contributions could focus on improving the performance of learning-based feature extraction to better compensate for background changes, similar objects, symmetry, and texture-less surfaces.

\section*{Acknowledgment}
The authors would like to express their sincere gratitude to Dr. Houman Masnavi for his invaluable contributions and support in the development of the ROS-based simulations.
% \newpage
\bibliography{ICUAS}
\bibliographystyle{ieeetr}
\end{document}